\documentclass[runningheads]{llncs}
\usepackage{graphicx}
\usepackage{makecell} 
\usepackage{multirow} 
\usepackage{subcaption}
\usepackage{booktabs} 
\usepackage{longtable} 
\usepackage{array} 
\usepackage{makecell} 
\usepackage{amsmath}
\usepackage{hyperref}
\usepackage{orcidlink}

\hypersetup{
  colorlinks = true,
  linkcolor =red,
  citecolor = green,
  urlcolor = blue
}

\newcolumntype{V}[1]{!{\vrule width #1}}
\newcolumntype{L}[1]{>{\raggedright\arraybackslash}m{#1}}
\newcolumntype{C}[1]{>{\centering\arraybackslash}m{#1}}
\newcolumntype{R}[1]{>{\raggedleft\arraybackslash}m{#1}}
\begin{document}
%
\title{ChartReformer: Natural Language-Driven Chart Image Editing}
%
%


\author{Pengyu Yan\orcidlink{0000-0003-1584-2350} \and
Mahesh Bhosale\orcidlink{0009-0003-7157-2129} \and
Jay Lal\orcidlink{0009-0000-0515-1287} \and 
Bikhyat Adhikari\orcidlink{2222--3333-4444-5555} \and 
David Doermann\orcidlink{0000-0003-1639-4561}}

\authorrunning{P. Yan et al.}



%
\institute{University at Buffalo, Buffalo, NY 14228, USA \\
\email{\{pyan4,mbhosale,jayashok,bikhyata,doermann\}@buffalo.edu}}
\maketitle              
\begin{abstract}
Chart visualizations are essential for data interpretation and communication; however, most charts are only accessible in image format and lack the corresponding data tables and supplementary information, making it difficult to alter their appearance for different scenarios of application. To eliminate the need for original underlying data and information to perform chart editing, we propose ChartReformer, a natural language-driven chart image editing solution that directly edits the charts from the input images with the given instruction prompts. Instead of predicting the plotting code, the key in this method is that we allow the model to comprehend the chart and reason over the prompt to generate the corresponding underlying data table and visual attributes for new charts, enabling a precise and stable editing result. To generalize ChartReformer, we define and standardize the chart editing category and generate the ChartCraft dataset, covering style, layout, format, and data-centric edits. The experiments show promising results for the natural language-driven chart image editing. Our datasets and model are available at: \url{https://github.com/pengyu965/ChartReformer}.

\end{abstract}
\section{Introduction}
Charts are designed with specific aesthetics and formats to effectively visualize tabular data. However, a given visualization may only be ideal for a specific scenario or purpose. Modifying chart images would allow them to be adapted for diverse applications by enabling the highlighting of specific data segments, amplifying the distinctions between data points, converting charts into different formats, or editing the appearance of the graph style. These are significant and can enhance accessibility for readers.\par 

In the evolving landscape of data analysis, charts and graphs play an indispensable role in deciphering complex datasets and facilitating informed decision-making. The ability to effectively visualize tabular data through charts is crucial, yet the specificity of a visualization's design to its initial context limits adaptability for broader applications. This necessitates the development of methods to modify chart images, enabling them to highlight particular data segments, enhance distinctions between data points, or improve accessibility for diverse readerships. However, traditional chart-editing methods are fraught with challenges. These processes often require significant manual intervention, a deep understanding of the plotting software's parameters, and access to the original data tables. These limitations become particularly acute in scenarios where source data are lost or unavailable, highlighting the need for more flexible and accessible editing techniques.\par 

\begin{figure}[tb]
    \centering
    \includegraphics[width=0.7\linewidth]{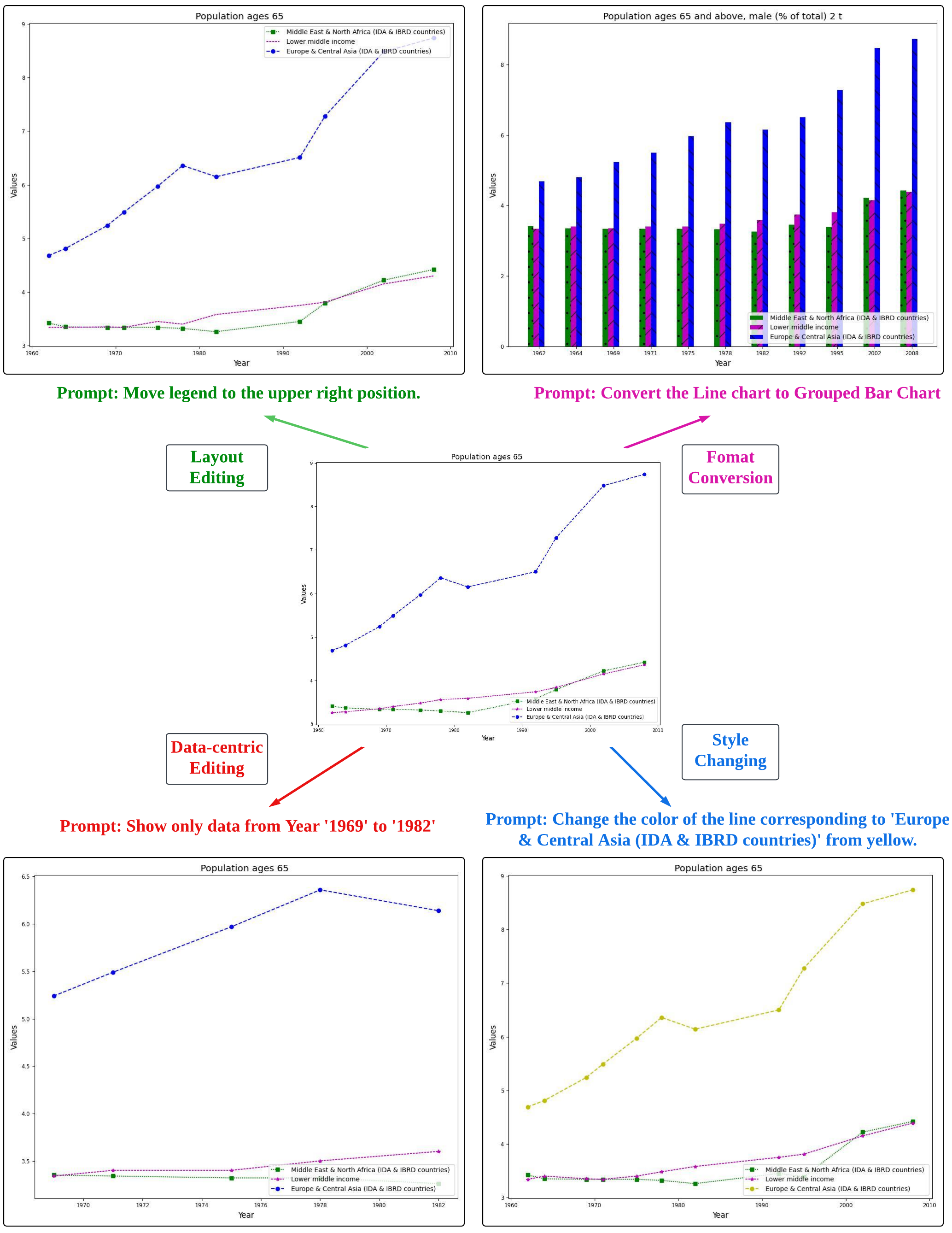}
    \caption{\textbf{Examples of chart editing results from our methods.} In total, our methods define and cover four types of chart editing: style, layout, format and data-centric edit.}
    \label{fig:chart_edit_illustration}
\end{figure}

Recent advances in computer vision and natural language processing (NLP) have opened new avenues for understanding charts. As multi-modal tasks, chart understanding related research topics such as data extraction, question answering, and chart summarization are tackled by utilizing visual language model in~\cite{liu2022matcha,liu2022deplot,han2023chartllama,masry2023unichart,cheng2023chartreader}. However, for chart editing tasks, ~\cite{liu2023improved} still rely on input visualization code and resource data table, while ChartLlama~\cite{han2023chartllama} fails to cover the full spectrum of possible edits, such as data manipulation for input charts. To close this gap, we introduce ChartReformer, which edits chart images from natural language prompts without any prior knowledge of the underlying data and original plot settings. Training on our dataset allows it to cover a wide range of edits from style, format, layout to data-centric edits. In our methods, we decompose the input charts and reasoning over the prompts for a new corresponding data table and visual attributes, allowing for detailed, comprehensive, and accurate chart editing. This approach predicting and adjusting the embedded visual attributes and data under original chart images enables the creation of customized chart images through a re-plotter without explicitly predicting the plotting code, producing and delivering robust and stable chart editing results. \par 

Overall the main contributions of our work can be summarized as follows: \par 

\begin{itemize}
    \item The first work thoroughly discusses the chart editing tasks. Define and standardize the editing category. Provide a detailed taxonomy of the edits. Suitable evaluation metrics are designed for such tasks.
    \item Provide ChartCraft datasets that span major edit categories, including style, format, layout, and data-centric edits. The datasets contain 100K pairs of original and edited chart images with corresponding underlying data tables, visual attributes, and instruction prompts. 
    \item Present ChartReformer pipeline with a visual language model trained on our dataset from deplot's checkpoint, and empirically demonstrate the effectiveness of our system in experiments.
\end{itemize}

\section{Related Work}
\subsection{Natural-language-driven visualization}
The intersection of Natural Language Processing (NLP) and data visualization within the field of human-computer interaction (HCI) has become increasingly prominent~\cite{8019860}. This surge in interest, especially within the deep learning community, is driven by advances in natural language understanding~\cite{nl4dv}. Tools such as VegaLite~\cite{vegalite} and ChartDialogs~\cite{chartdialogs} demonstrate the ability to generate and adjust visual charts in natural language, the former utilizing JSON for chart specifications and the latter applying Seq2Seq models for editing in natural language dialogues. Similarly, VizGpt(\url{vizgpt.ai}) employs GPT models to react with human language instructions for visualization and styling. These approaches are practical, yet different from ours; they rely on available underlying tabular data and do not directly alter visualized images.

\subsection{Chart Comprehension}

\subsubsection{Datasets} 
Editing charts is a nuanced task that necessitates a grasp of both the chart's visual features and the data it represents. Contemporary multimodal models such as GPT-4V~\cite{yang2023dawn} and LLaVA-1.5 face challenges in analyzing and extracting the underlying data of the charts~\cite{guan2023hallusionbench}, while chart manipulation is even more difficult. To facilitate the model in understanding the charts, several datasets have been introduced. Some assess understanding via straightforward question-and-answer formats with human-annotated QA pairs - for example, ChartQA~\cite{masry-etal-2022-chartqa}, or utilize templates from crowd-sourced platforms - like PlotQA~\cite{plotqa}, or employ synthetic examples created by Large Language Models (LLMs) - as seen in ChartLlama~\cite{han2023chartllama}. Other dataset categories, such as chart-to-text~\cite{chart-to-text}, measure comprehension through summarization. However, to our knowledge, there is no publicly available datasets for chart editing.\par 

\subsubsection{Models}
Many existing methods~\cite{lal2023lineformer,luo2021chartocr,yan2023chartdete,ahmed2023spaden} analyze the component of the charts and extract the underlying data by relating the results of component detection. This routine relies heavily on the intermediate results and is potentially vulnerable. Utilizing the large visual language model, direct understanding and reasoning can be performed on input charts without the need for intermediate steps. Pix2Struct~\cite{pix2struct} is a such model on similiar task that extracts visually-situated language from web page screenshot into structured text. Then Matcha~\cite{liu2022matcha} adapts Pix2Struct for chart reasoning by pre-training the model on plot
deconstruction and numerical reasoning. Deplot~\cite{liu2022deplot} fine-tune the matcha for chart-to-table conversion, while this conversion will loss the crucial apperance references, vital for chart editing. Both Matcha~\cite{liu2022matcha} and ChartLlama~\cite{han2023chartllama} can de-render chart images into tables and plot code. However, none of these models is explicitly capable of predicting and adjusting the charts' visual attributes and corresponding underlying data, which is vital to editing charts accurately. Meanwhile, predicting code is a challenging way to deliver the final edit results, since they have a higher chance of failure in code compilation. 

\section{Problem Statement}

\subsection{Chart Edit Taxonomy}
Surveying the typical edits performed on chart images, we can categorize them into four distinct classes: style, layout, format, and data-centric, as detailed in Table~\ref{tab:edit_spec}. Furthermore, most advance chart edits can be obtained by chaining these individual edits. The following subsections elaborate on each of the different edit categories. \par 

\begin{table}[tb]
    \centering
    \caption{Chart Edit Taxonomy}
    \begin{tabular}
    {C{6em}|C{10em}|L{16em}}
    \Xhline{1.5pt}
    \textbf{Edit Category} & \textbf{Aspect} & \textbf{Attributes} \\
    \Xhline{1pt}
        \multirow{4}{*}{Style} & \multirow{1}{*}{Colors} & Plot Color   \\
         \cline{2-3}
         & Line & Line Style, Marker \\
         \cline{2-3}
         & Pattern & Bar Pattern  \\
         \cline{2-3}
         & \multirow{1}{*}{Text} & Font Family, Font Size \\
    \Xhline{1pt}
        \multirow{2}{*}{Layout}
         & \multirow{1}{*}{Axes} & Grid Lines \\ 
         \cline{2-3}
         & \multirow{1}{*}{Legend}  & Position,  Internal Layout \\
    \Xhline{1pt}
        \multirow{3}{*}{Format} & \multirow{3}{*}{Plot} & Line $\leftrightarrow$ Grouped  Bar \\
         & & Line $\leftrightarrow$ Stacked Bar \\
         & & Grouped Bar $\leftrightarrow$ Stacked Bar \\
    \Xhline{1pt}
         \multirow{4}{*}{Data Centric}& \multirow{2}{*}{Data Filtering} & Range-based \\
         & & Series-based \\ 
         \cline{2-3}
         & \multirow{1}{*}{Data Addition} & Add/Update Data-point, Add/Update Data-series \\
    \Xhline{1pt}
    \end{tabular}
    \vspace{0.5em}
    \label{tab:edit_spec}
\end{table}

\subsubsection{Style Edits}: Style in chart images encompasses essential low-level features such as plot colors, styles (including line style, marker style, and bar pattern), and font characteristics (type, size, etc.). Modifications to a chart's style strictly alter these foundational visual attributes of individual elements without tampering with the data or visualization format. Such appearance-based tweaks are crucial for various applications. Considering accessibility: A color-differentiated chart may not be discernible to color-blind individuals, and transitioning to a different color palette can significantly enhance the chart comprehension for them. \par 

\subsubsection{Layout Re-composition}: Our layout considerations encompass two primary aspects: axes grids and legends. Tweaks layouts in chart editing can ensure that each element is represented systematically in the modified chart. Furthermore, specific layout changes can improve data readability, such as the inclusion or exclusion of grid lines.\par 

\subsubsection{Format Conversion}: Different chart types highlight specific aspects of the data. Line charts are particularly effective at showcasing trends over time, allowing viewers to quickly discern patterns and changes. Bar charts, on the other hand, excel in comparing quantities among different groups or categories. More specifically, grouped bar charts show mostly the absolute value comparison, while stacked bar charts reflect the ratio. Transitioning between these formats can offer more comprehensive data views from different aspects.\par 

\subsubsection{Data-Centric Modifications}: Data-centric chart modifications enable precise manipulation of the chart's underlying data, facilitating tailored adjustments to the visualization's displayed information. These edits, which range from subtle alterations to significant additions, can dramatically shift the narrative and insights gleaned from a chart. Critical operations include range filtering to focus on specific data segments, series filtering to highlight particular data series, and adding new data series or data points to enrich the visualization. Such direct interactions with the chart data empower users to create custom views and explore data in novel and informative ways.

\section{ChartReformer}

\begin{figure}[tb]
    \centering
    \includegraphics[width=\linewidth]{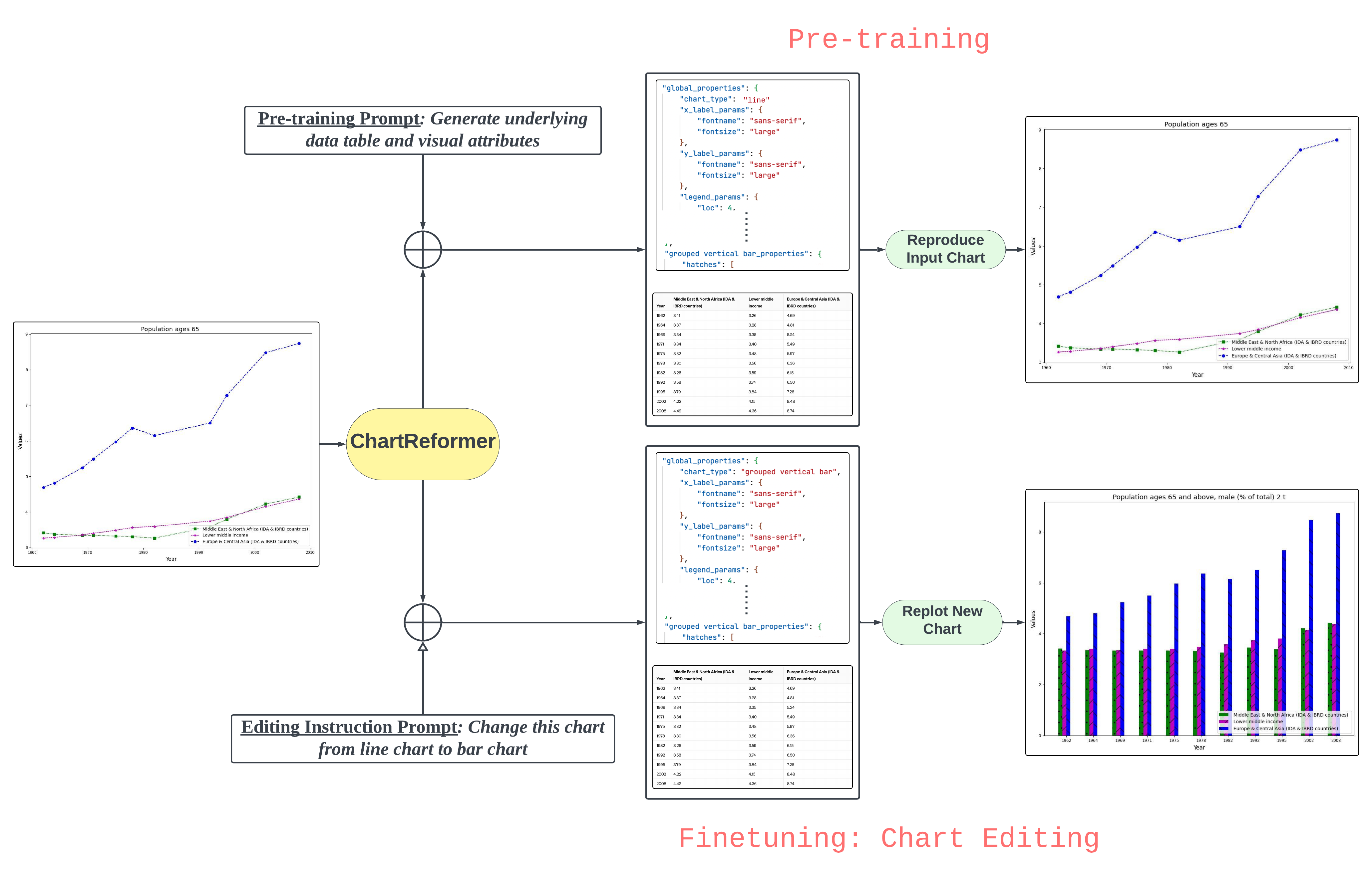}
    \caption[ChartReformer Method]{A Chart Image and edit-prompt are taken as input by the ChartReformer model, which predicts visual attributes and data for the corresponding edited chart. A Replotter software takes in these predicted parameters and generates the edited chart-image}
    \label{fig:chartreformer_method}
\end{figure}

Chart images are inherently complex. Hence, accurate chart comprehension is a prerequisite to successful chart editing. The structural decomposition of the chart image has been seen as a potential approach to the problem. Matcha~\cite{liu2022matcha} and ChartLlama~\cite{han2023chartllama} try to predict the Python visualization code corresponding to the chart. However, predicting accurate plotting code for the wide variety of charts proves to be a challenging task and easy to fail in success code compilation. We propose ChartReformer, a method that de-renders charts into underlying data and visual attributes to address these limitations for chart editing. Fig.~\ref{fig:chartreformer_method} shows how ChartReformer predicts a decomposed chart image that a replotter software can effectively utilize to construct the edited charts. To allow the model to reason over the prompt accurately generating the visual attributes and underlying data, a dataset aligning our pipeline is required. The following sections introduce our dataset and pipeline in detail. \par 

\subsection{Dataset}
This section provides insight into the data creation process for Chart Edits. We synthesize paired chart images using data tables from existing chart datasets to obtain a sizeable chart editing dataset. Our primary emphasis is on line and bar charts, including both grouped and stacked vertical/horizontal bar variations, given their prevalent use in real-world charting scenarios.

\subsubsection{Data Source}: We utilized data tables from AdobeSynth-19 data previously released as part of the ICPR ChartInfographics competition~\cite{icpr2020comp}. This dataset originally consists of synthetically generated images using Matplotlib~\cite{matplotlib}, however, the underlying data is derived from real-world sources such as World Development Indicators and Gender Statistics (World Bank), among others.

\subsubsection{Chart Image Generation}:
We developed a custom software using Matplotlib to synthesize chart images with varying visual attributes given as input. The tool supports all the edits specified in Table~\ref{tab:edit_spec} while allowing sampling from a comprehensive pool of visual attributes. A thorough parameter pool is provided in the Appendix Table~\ref{Table: Matplotlib property pool}. Parameters are randomly selected from this pool, a strategy that introduces significant diversity in the visual appearance of the generated charts. Since plotting parameters are explicitly defined, storing them in a modifiable JSON format becomes straightforward, facilitating further adjustments and reuse. Appendix Fig.~\ref{fig:json_example} shows an example for such a JSON specifying all parameters.

\subsubsection{Edit Pair Synthesis}:
The chart editing data generation software produces, for every sample, edited chart pairs and corresponding edit text prompts. Each chart in the pair consists of visual attributes, a data table, and the synthesized image. The details of generating pairs for each editing category are described as follows:

\begin{itemize}
    \item \textbf{Style and Layout Edits:} For each style or layout modification outlined in the edit specifications, we modify the relevant plotting parameter in the parameter JSON file.
    \item \textbf{Format Edits:} This edit category facilitates the conversion between line and bar charts, and vice versa. We employ the identical data table, generating a new chart type and substituting the original plotting parameters with those corresponding to the new chart, the color of each data series before and after conversion remains the same, while we allow the style like hatches or markers for new plot to be different, e.g., from line chart to bar chart, the line color and bar color are corresponding to each other while bar pattern can be randomly chosen.
    \item \textbf{Data Edits:} Here, we alter the data table according to the given prompt and tailor the original plotting parameters accordingly. This ensures that while the visual attributes of the chart remain consistent, the data in the charts are updated.    
\end{itemize}\par 

For each pair of edits, 5-7 varied prompts are generated from one base prompt to help the model capture the diversity of natural language expressions. \par 

\subsubsection{Dataset Statistics}: 
Table~\ref{tab:dataset_statistics} outlines the statistics of our datasets, categorizing them into four principal types of edits. Overall, our dataset encompasses approximately 100k paired samples. Data-centric edit has significantly more number of samples due to the task challenges. \par 

\begin{table}[ht]
\centering
\caption{Dataset Statistics: Number of paired samples by Edit Types across Chart Categories}
\label{tab:dataset_statistics}
\begin{tabular}{|l|c|c|c|c|c|c|} 
\hline
\textbf{Chart Type} & \textbf{Style} & \textbf{Layout} & \textbf{Format} & \textbf{Data-centric} & \textbf{Total} \\
\hline
Line         & 4775 & 1910 & 3820 & 11460 & 21965 \\
Grouped Vertical Bar        & 4775 & 1910 & 3820 & 11460 & 21965 \\
Grouped Horizontal Bar  & 4775 & 1910 & 3820 & 11460 & 21965 \\
Stacked Vertical Bar        & 4775 & 1910 & 3820 & 11460 & 21965 \\
Stacked Horizontal Bar & 4775 & 1910 & 3820 & 11460 & 21965 \\
\hline
Total & 23875 & 9550 & 19100 & 57300 & 109825 \\
\hline
\end{tabular}
\end{table}

\begin{figure}[t]
    \centering
    \includegraphics[width=\linewidth]{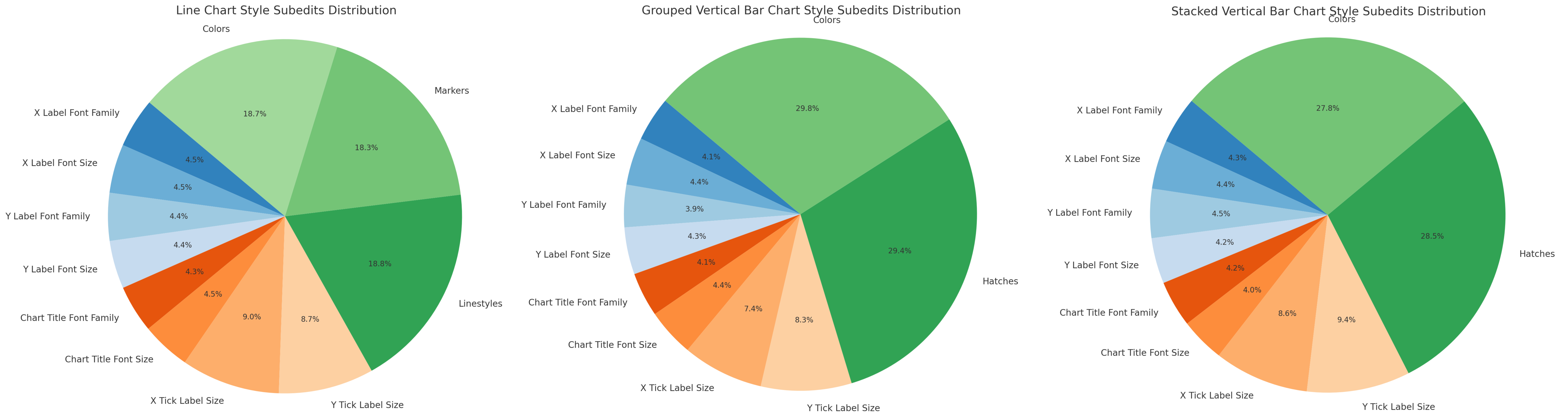}
    \caption{Distribution of Samples for Style Edits across Chart Categories}
    \label{fig:style_subedit_distribution_subfig}
\end{figure}

The distribution of detailed edits for style, layout, format and data-centric editing is shown in Fig,\ref{fig:style_subedit_distribution_subfig}. The generation of chart-editing dataset is based on the manipulation of visual attribute parameters. There are overall two types of visual attribute parameters, chart-type-relevant parameters (line styles/markers, bar patterns, etc.), and chart-type-irrelevant parameters (font size, axis labels orientation, etc.). The edits based on the first type of parameters are oversampled than the second type, since it helps model to learn more challenging part -- identifying and manipulating the plotting graph. \par 

\subsection{Our Pipeline}

To address the challenge of chart editing, we adapt the visual-language encoder-decoder-based transformer~\cite{pix2struct} to our chart editing tasks. We break down the training into two stages: pre-training for accurate chart de-rendering and fine-tuning for chart editing.\par 

\subsubsection{Chart De-rendering}
Accurately de-rendering the chart to visual attributes and underlying data is a prerequisite. We pre-train ChartReformer on our dataset with unpaired images to enable accurate visual attributes and underlying data extraction. Simultaneously predicting the underlying data and visual attributes allows the model to learn the mapping between them. We initialize the model with the checkpoint from~\cite{liu2022deplot} and train it on our dataset with 100K samples (sampled from each side of the paired charts). To avoid text distortion and blur during the input resizing of chart images, we opt for a larger input image size of $(800,800)$ with padding to maintain the original aspect ratio. The maximum output sequence length is 1024, allowing sufficient prediction of all parameters and data tables. \par 

\subsubsection{Chart Editing}

We fine-tune the pretrained model on paired images and edit prompts, totally 88k samples, enabling it to interpret prompts and adjust data and visual properties accordingly. Edited charts are then replotted with predicted data and plotting parameters. To enhance real-world plotting success, we suggest using JSON repair and default plotting parameters for incomplete predictions, though our evaluations eschew repairs for unbiased performance assessment.

\section{Experiments}

We use ChartLlama as a baseline for comparison. To the best of our knowledge, no existing dataset related to chart editing is publicly available\footnote{Based on current information, the dataset developed by ChartLlama~\cite{han2023chartllama} has not yet been formally published}. Hence, we perform an evaluation exclusively on our dataset.

\subsection{Metrics}

\subsubsection{Image-based Evaluation}
To facilitate a model-agnostic comparison, we utilize Structural Similarity Index Measure (SSIM)~\cite{ssim} to assess the quality of the generated image relative to the edited image from the ground truth. SSIM offers a nuanced perspective on the degree to which the edited chart mirrors the expected outcome, capturing subtle and critical edits speaking to structural similarity aspective. We also calculate a success rate, which reflects the proportion of edits where the edited image was successfully generated, accounting for instances of plotting failures due to inaccurate or incomplete structure prediction. For comparison method ChartLlama, the success rate measures the ratio of samples that the predicted code can be successfully compiled.\par 

\begin{table}[tb]
    \centering
    \caption{ChartReformer performance across different types of edits. The first two rows represent Visual Attributes and Data-Table scores for edits, whereas the last two represent image-level comparison metrics}
    \resizebox{\linewidth}{!}{
    \begin{tabular}{ c | c c c | c c c | c c c | c c c | c c c }
    \Xhline{1.5pt}
        & \multicolumn{3}{c|}{Style} & \multicolumn{3}{c|}{Layout} & \multicolumn{3}{c|}{Format} & \multicolumn{3}{c|}{Data-Centric} & \multicolumn{3}{c}{Total} \\
    \cline{2-16}
         Metrics & P & R & F1 & P & R & F1 & P & R & F1 & P & R & F1 & P & R & F1 \\
    \Xhline{1pt}
         VAES & 86.64 & 86.62 & 86.63 & 89.66 & 89.64 & 89.65 & 90.54 & 90.52 & 90.53 & 84.25 & 84.22 & 84.24 & 86.32 & 86.31 & 86.32\\ 
         RMS & 90.45 & 90.28 & 90.35 & 89.33 & 89.16 & 89.23 & 91.27 & 91.18 & 91.22 & 88.45 & 88.48 & 88.36 & 89.42 & 89.37 & 89.33\\
    \cline{1-16}
         SSIM & \multicolumn{3}{c|}{84.5} & \multicolumn{3}{c|}{82.06} & \multicolumn{3}{c|}{85.19} & \multicolumn{3}{c|}{83.07} & \multicolumn{3}{c}{83.65} \\
         Success Rate & \multicolumn{3}{c|}{99.81} & \multicolumn{3}{c|}{100} & \multicolumn{3}{c|}{99.9} & \multicolumn{3}{c|}{99.77} & \multicolumn{3}{c}{99.9} \\
    \Xhline{1pt}
    \end{tabular}}
    \label{evaluation}
\end{table}

\subsubsection{Evaluating Edit Correctness}
To offer a more precise evaluation of the performance of our methods on the chart editing task, we use the predicted visual attributes and underlying data table for evaluation. Our replotting process is heurstic, therefore, comparing those with ground truth reflect the performance well. We use the Relative Mapping Similarity (RMS) from~\cite{liu2022deplot}, and Visual Attribute Edit score (VAES) to evaluate the accuracy of underlyting data table and visual attributes prediction, respectively.\par 

\begin{equation}
\begin{aligned}
     S_{changed} &= \frac{1}{|X_c|} \sum_{e \in X_c} S(e, g) \\
     S_{unchanged} &= \frac{1}{|X_u|} \sum_{e \in X_u} S(e, g)
\end{aligned}
\label{VAES}
\end{equation}

\begin{equation}
S_f = \frac{2 \cdot S_{changed} \cdot S_{unchanged}}{S_{changed} + S_{unchanged}}
\label{VAES_aggre}
\end{equation}

VAES is calculated by grouping attributes into two groups: attributes should be edited $S_{changed}$, and attributes should remain unchanged $S_{unchanged}$, as shown in Equation~\ref{VAES}. $e$ and $g$ represent the edited attribute value and the corresponding ground truth value. First, we calculate a similarity matrix to match the key/value between the ground truth and the prediction. Then the score for each key is calculated based on the value type: the categorical value are based on an exact match; and numeric value are scores from 0 to 1 based on threshold of 0.4. Finally, the visual attribute edit score (VAES) $S_f$ is calculated using the harmonic mean among them as Equation~\ref{VAES_aggre}. This prevents biased evaluation, as only a minor part of the visual attributes will change. The precision, recall and F1 score are obtained based on the matching result.\par 

\begin{table}[htb]
    \centering
    \caption{Comparison with ChartLlama across different edits on a subset of the test-set (550 samples, 10\% sub-test samples)}
    \resizebox{\linewidth}{!}{
    \begin{tabular}{c|c c|c c|c c|c c|c c}
    \Xhline{1.5pt}
         & \multicolumn{2}{c|}{Style} & \multicolumn{2}{c|}{Layout} & \multicolumn{2}{c|}{Format} & \multicolumn{2}{c|}{Data-Centric} & \multicolumn{2}{c}{Total} \\
         & SSIM & Success Rate & SSIM & Success Rate & SSIM & Success Rate & SSIM & Success Rate & SSIM & Success Rate \\
    \Xhline{1pt}
        ChartLlama & 73.09 & 10.63 & 64.77 & 26.47 & 64.32 & 8.04 & 67.68 & 19.33 & 67.46 & 16.95 \\ 
    \hline
        \textbf{ChartReformer} & 83.3 & 100 & 82.55 & 100 & 84.22 & 100 & 81.43 & 100 & 82.39 & 100 \\
    \Xhline{1.5pt}
    \end{tabular}}
    \label{tab:comparison_chartllama}
\end{table}


\subsection{Results}

\begin{figure}
    \centering
    \includegraphics[width=0.9\linewidth]{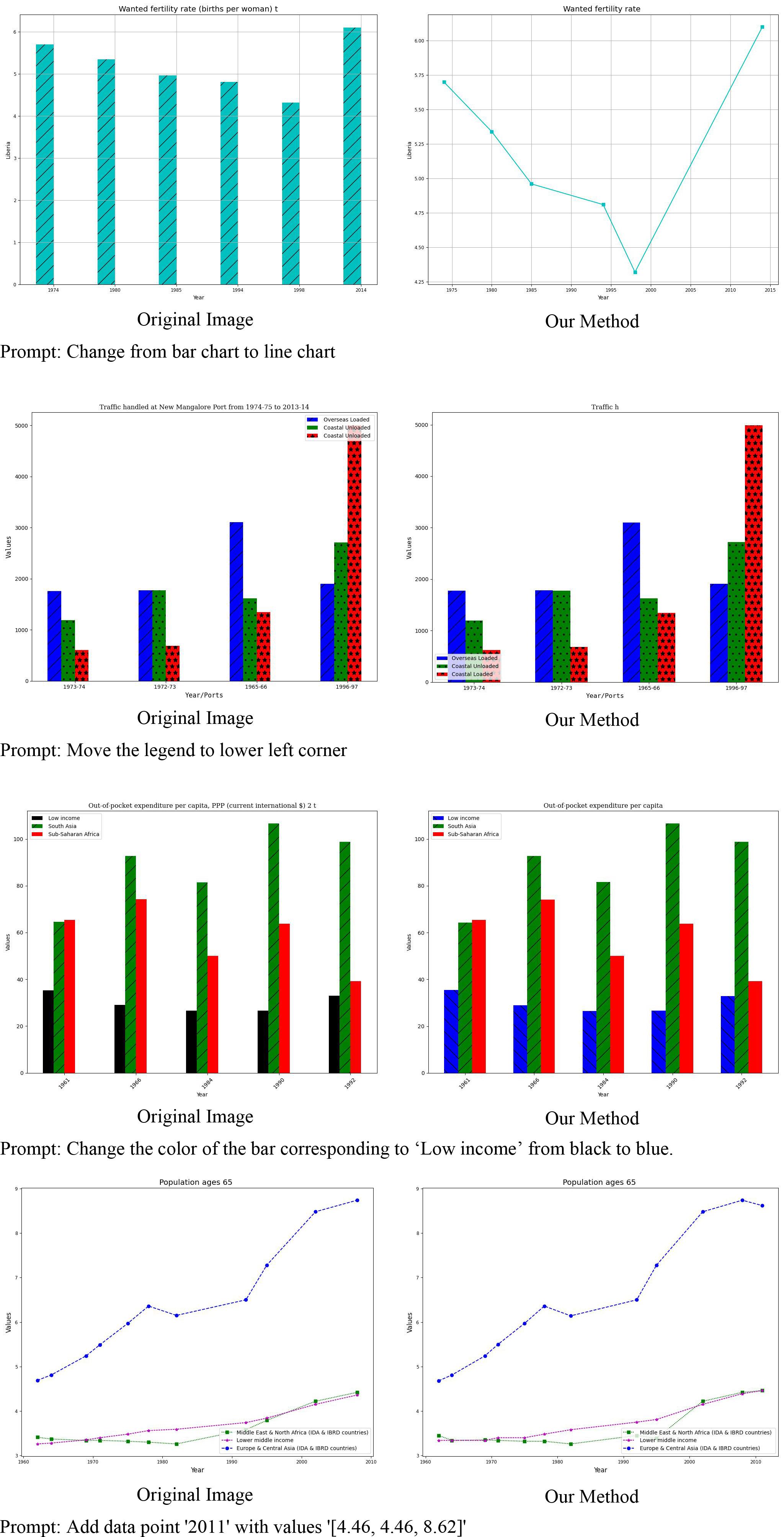}
    \caption{Qualitative Results for ChartReformer}
    \label{fig:chartformer_qualitative_eval}
\end{figure}

Table~\ref{evaluation} shows the results of ChartReformer on our test set consisting of 5.5K samples across different edit types. VAES and RMS measure edit-correctness corresponding to visual attributes and data, respectively, whereas SSIM measures image-level similarity with the ground-truth edited image. The results show that data-centric edits appear to be the hardest as they require a precise understanding of the existing data and manipulation based on the prompt while the format editing is easier since it requires least change in the visual attributes json based on the method setup. Overall, the model performs reasonably well on chart edits while maintaining high fidelity with the original chart image, as also seen in Fig.~\ref{fig:chartformer_qualitative_eval}.\par 

Compared to recent work ChartLlama~\cite{han2023chartllama}, since ChartLlama follows a different methodolgy than ours, we could only compare with ChartLlama with SSIM and success rate. From Table~\ref{tab:comparison_chartllama}, ChartReformer performs better across all edit categories. This performance gap could be attributed to the lack of training on a comprehensive edit dataset. Further, the success rate of our method is better as we do not predict the visualization python code, which is harder to get correct. We concede that in the current setup, we did not perform prompt engineering for ChartLlama, which will likely drag the performance of ChartLlama.\par

\begin{figure}
    \centering
    \includegraphics[width=\linewidth]{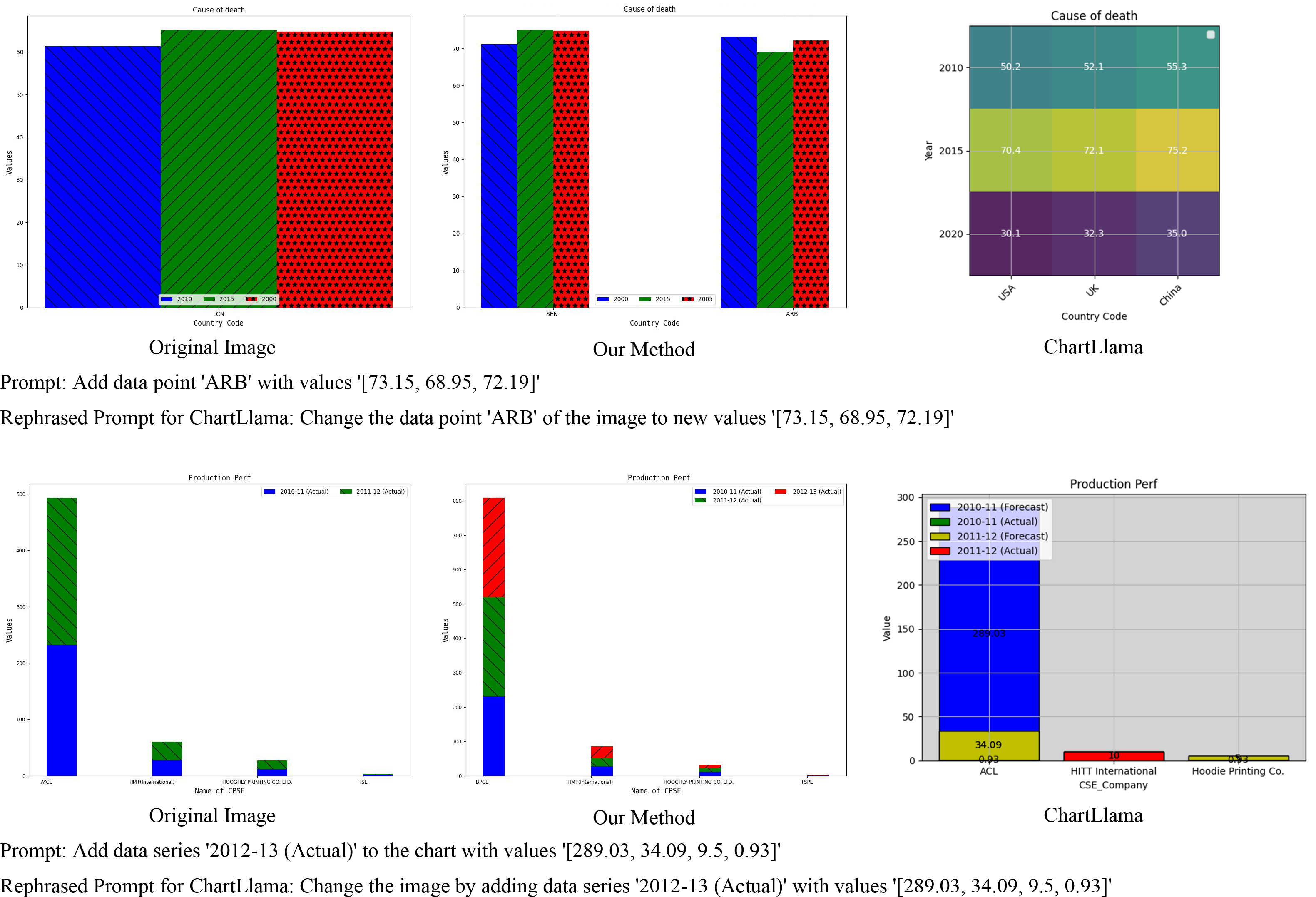}
    \caption{Qualitative Results for ChartLlama}
    \label{fig:chartllama_qualitative_eval}
\end{figure}

\section{Discussion and Limitations}
As shown in Table~\ref{evaluation}, ChartReformer can successfully extract and alternate visual attributes accordingly for all types of chart edits. In the experiments, we noticed that the overall edit performance is heavily dependent on the data extraction accuracy. Therefore, a more accurate data extraction approach would result in more precise data editing and would be a promising avenue for further research. The proposed chart-edit dataset covers a wide range of edits, yet real-world edit instructions could be abstract and arbitrarily complex, e.g., 'Modify the chart color palette to make it accessible to colorblind people'. One way to handle such queries is to use a preprocessing module (for instance, a decoder-only language model), which could be used to better interpret and simplify them into a series of simply chained edit prompts that ChartReformer can handle. 

\section{Conclusion}
In this work, we present and standardize the chart editing task and generate a large dataset, namely ChartCraft. ChartReformer presents a novel approach to chart editing, allowing for modifications directly from chart images without the need for underlying data tables or supplementary information. By generating edited charts in a decomposed form that includes both the data table and visual attributes, ChartReformer enables precise, natural language-driven edits across style, layout, format, and data-centric modifications. Our experiments demonstrate promising results, highlighting ChartReformer's potential to enhance chart accessibility and adaptability for diverse applications.

\bibliographystyle{splncs04}
\bibliography{references}

\begin{thebibliography}{10}
\providecommand{\url}[1]{\texttt{#1}}
\providecommand{\urlprefix}{URL }
\providecommand{\doi}[1]{https://doi.org/#1}

\bibitem{ahmed2023spaden}
Ahmed, S., Yan, P., Doermann, D., Setlur, S., Govindaraju, V.: Spaden: Sparse and dense keypoint estimation for real-world chart understanding. In: International Conference on Document Analysis and Recognition. pp. 77--93. Springer (2023)

\bibitem{cheng2023chartreader}
Cheng, Z.Q., Dai, Q., Li, S., Sun, J., Mitamura, T., Hauptmann, A.G.: Chartreader: A unified framework for chart derendering and comprehension without heuristic rules. arXiv preprint arXiv:2304.02173  (2023)

\bibitem{icpr2020comp}
Davila, K., Tensmeyer, C., Shekhar, S., Singh, H., Setlur, S., Govindaraju, V.: Icpr 2020 - competition on harvesting raw tables from infographics. In: Del~Bimbo, A., Cucchiara, R., Sclaroff, S., Farinella, G.M., Mei, T., Bertini, M., Escalante, H.J., Vezzani, R. (eds.) Pattern Recognition. ICPR International Workshops and Challenges. pp. 361--380. Springer International Publishing, Cham (2021)

\bibitem{guan2023hallusionbench}
Guan, T., Liu, F., Wu, X., Xian, R., Li, Z., Liu, X., Wang, X., Chen, L., Huang, F., Yacoob, Y., Manocha, D., Zhou, T.: Hallusionbench: An advanced diagnostic suite for entangled language hallucination \& visual illusion in large vision-language models (2023)

\bibitem{han2023chartllama}
Han, Y., Zhang, C., Chen, X., Yang, X., Wang, Z., Yu, G., Fu, B., Zhang, H.: Chartllama: A multimodal llm for chart understanding and generation. arXiv preprint arXiv:2311.16483  (2023)

\bibitem{matplotlib}
Hunter, J.D.: Matplotlib: A 2d graphics environment. Computing in Science \& Engineering  \textbf{9}(3),  90--95 (2007). \doi{10.1109/MCSE.2007.55}

\bibitem{chart-to-text}
Kantharaj, S., Leong, R.T., Lin, X., Masry, A., Thakkar, M., Hoque, E., Joty, S.: Chart-to-text: A large-scale benchmark for chart summarization. In: Muresan, S., Nakov, P., Villavicencio, A. (eds.) Proceedings of the 60th Annual Meeting of the Association for Computational Linguistics (Volume 1: Long Papers). pp. 4005--4023. Association for Computational Linguistics, Dublin, Ireland (May 2022). \doi{10.18653/v1/2022.acl-long.277}, \url{https://aclanthology.org/2022.acl-long.277}

\bibitem{lal2023lineformer}
Lal, J., Mitkari, A., Bhosale, M., Doermann, D.: Lineformer: Line chart data extraction using instance segmentation. In: International Conference on Document Analysis and Recognition. pp. 387--400. Springer (2023)

\bibitem{pix2struct}
Lee, K., Joshi, M., Turc, I., Hu, H., Liu, F., Eisenschlos, J., Khandelwal, U., Shaw, P., Chang, M.W., Toutanova, K.: Pix2struct: screenshot parsing as pretraining for visual language understanding. In: Proceedings of the 40th International Conference on Machine Learning. ICML'23, JMLR.org (2023)

\bibitem{liu2022deplot}
Liu, F., Eisenschlos, J.M., Piccinno, F., Krichene, S., Pang, C., Lee, K., Joshi, M., Chen, W., Collier, N., Altun, Y.: Deplot: One-shot visual language reasoning by plot-to-table translation. arXiv preprint arXiv:2212.10505  (2022)

\bibitem{liu2022matcha}
Liu, F., Piccinno, F., Krichene, S., Pang, C., Lee, K., Joshi, M., Altun, Y., Collier, N., Eisenschlos, J.M.: Matcha: Enhancing visual language pretraining with math reasoning and chart derendering. arXiv preprint arXiv:2212.09662  (2022)

\bibitem{liu2023improved}
Liu, H., Li, C., Li, Y., Lee, Y.J.: Improved baselines with visual instruction tuning (2023)

\bibitem{luo2021chartocr}
Luo, J., Li, Z., Wang, J., Lin, C.Y.: Chartocr: Data extraction from charts images via a deep hybrid framework. In: Proceedings of the IEEE/CVF winter conference on applications of computer vision. pp. 1917--1925 (2021)

\bibitem{masry-etal-2022-chartqa}
Masry, A., Do, X.L., Tan, J.Q., Joty, S., Hoque, E.: {C}hart{QA}: A benchmark for question answering about charts with visual and logical reasoning. In: Muresan, S., Nakov, P., Villavicencio, A. (eds.) Findings of the Association for Computational Linguistics: ACL 2022. pp. 2263--2279. Association for Computational Linguistics, Dublin, Ireland (May 2022). \doi{10.18653/v1/2022.findings-acl.177}, \url{https://aclanthology.org/2022.findings-acl.177}

\bibitem{masry2023unichart}
Masry, A., Kavehzadeh, P., Do, X.L., Hoque, E., Joty, S.: Unichart: A universal vision-language pretrained model for chart comprehension and reasoning. arXiv preprint arXiv:2305.14761  (2023)

\bibitem{plotqa}
Methani, N., Ganguly, P., Khapra, M.M., Kumar, P.: Plotqa: Reasoning over scientific plots. In: 2020 IEEE Winter Conference on Applications of Computer Vision (WACV). pp. 1516--1525 (2020). \doi{10.1109/WACV45572.2020.9093523}

\bibitem{nl4dv}
{Narechania}, A., {Srinivasan}, A., {Stasko}, J.: {NL4DV}: A {Toolkit} for generating {Analytic Specifications} for {Data Visualization} from {Natural Language} queries. IEEE Transactions on Visualization and Computer Graphics (TVCG)  (2020). \doi{10.1109/TVCG.2020.3030378}

\bibitem{vegalite}
Satyanarayan, A., Moritz, D., Wongsuphasawat, K., Heer, J.: Vega-lite: A grammar of interactive graphics. IEEE Transactions on Visualization and Computer Graphics  \textbf{23}(1),  341–350 (jan 2017). \doi{10.1109/TVCG.2016.2599030}, \url{https://doi.org/10.1109/TVCG.2016.2599030}

\bibitem{chartdialogs}
Shao, Y., Nakashole, N.: {C}hart{D}ialogs: {P}lotting from {N}atural {L}anguage {I}nstructions. In: Proceedings of the 58th Annual Meeting of the Association for Computational Linguistics. pp. 3559--3574. Association for Computational Linguistics, Online (Jul 2020). \doi{10.18653/v1/2020.acl-main.328}, \url{https://aclanthology.org/2020.acl-main.328}

\bibitem{8019860}
Srinivasan, A., Stasko, J.: Orko: Facilitating multimodal interaction for visual exploration and analysis of networks. IEEE Transactions on Visualization and Computer Graphics  \textbf{24}(1),  511--521 (2018). \doi{10.1109/TVCG.2017.2745219}

\bibitem{ssim}
Wang, Z., Bovik, A., Sheikh, H., Simoncelli, E.: Image quality assessment: from error visibility to structural similarity. IEEE Transactions on Image Processing  \textbf{13}(4),  600--612 (2004). \doi{10.1109/TIP.2003.819861}

\bibitem{yan2023chartdete}
Yan, P., Ahmed, S., Doermann, D.: Context-aware chart element detection. In: Fink, G.A., Jain, R., Kise, K., Zanibbi, R. (eds.) Document Analysis and Recognition - ICDAR 2023. pp. 218--233. Springer Nature Switzerland, Cham (2023)

\bibitem{yang2023dawn}
Yang, Z., Li, L., Lin, K., Wang, J., Lin, C.C., Liu, Z., Wang, L.: The dawn of lmms: Preliminary explorations with gpt-4v(ision) (2023)

\end{thebibliography}

\newpage
\section{Appendix}
\subsection{Dataset}
\subsubsection{Matplotlib visual attributes pool}: Table~\ref{Table: Matplotlib property pool} describes the Matplotlib parameters that we randomly sample from to generate our dataset have more diverse charts instead of relying on Matplotlib's default selection of these properties.

\begin{longtable}{>{\raggedright\arraybackslash}p{1.5cm} >{\raggedright\arraybackslash}p{2.5cm} >{\raggedright\arraybackslash}p{5cm} >{\centering\arraybackslash}p{1.5cm}}
\caption{Matplotlib Property Pool} \\
\toprule
\textbf{Scope} & \textbf{Property} & \textbf{Pool} & \textbf{Editable} \\
\midrule
\endhead
Line & Color & [``b'', ``g'', ``r'', ``c'', ``m'', ``y'', ``k''] & Yes \\
Line & Marker & [``o'', ``\verb|^|'', ``s'', ``*'', ``None''] & Yes \\
Line & Line-style & [``solid'', ``dashed'', ``dotted'', ``dense dotted'', ``loose dotted'', ``dense dashed'', ``loose dashed''] & Yes \\
Bar & Hatch & [``xx'', ``.'', ``*'', ``/'', ``\textbackslash'', ``None''] & Yes \\
Bar & Color & [``b'', ``g'', ``r'', ``c'', ``m'', ``y'', ``k''] & Yes \\
Global & X-axis Label Font Name & [``monospace'', ``Serif'', ``sans-serif'', ``Arial Black''] & Yes \\
Global & X-axis Label Font Size & [``medium'', ``large'', ``x-large''] & Yes \\
Global & Y-axis Label Font Name & [``monospace'', ``Serif'', ``sans-serif'', ``Arial Black''] & Yes \\
Global & Y-axis Label Font Size & [``medium'', ``large'', ``x-large''] & Yes \\
Global & Legend Location & [0, 1, 2, 3, 4, 8, 9] & Yes \\
Global & Legend Columns & [1, 2, 3] & No \\
Global & Title Font Name & [``monospace'', ``Serif'', ``sans-serif'', ``Arial Black''] & Yes \\
Global & Title Font Size & [``medium'', ``large'', ``x-large''] & Yes \\
Global & X-tick Label Size & [``x-small'', ``small'', ``medium'', ``large''] & Yes \\
Global & X-tick Rotation & [0, 45] & No \\
Global & Y-tick Label Size & [``x-small'', ``small'', ``medium'', ``large''] & Yes \\
Global & Grid Visibility & [True, False] & Yes \\
Global & Grid Axis & [`both', `x', `y'] & No \\
Global & Grid Line-style & [`solid', `dashed'] & No \\
\bottomrule
\label{Table: Matplotlib property pool}
\end{longtable}

\subsubsection{Example of JSON Properties}

\begin{figure}[ht]
    \centering
    \begin{minipage}{0.5\textwidth}
    \centering
    \includegraphics[width=\textwidth]{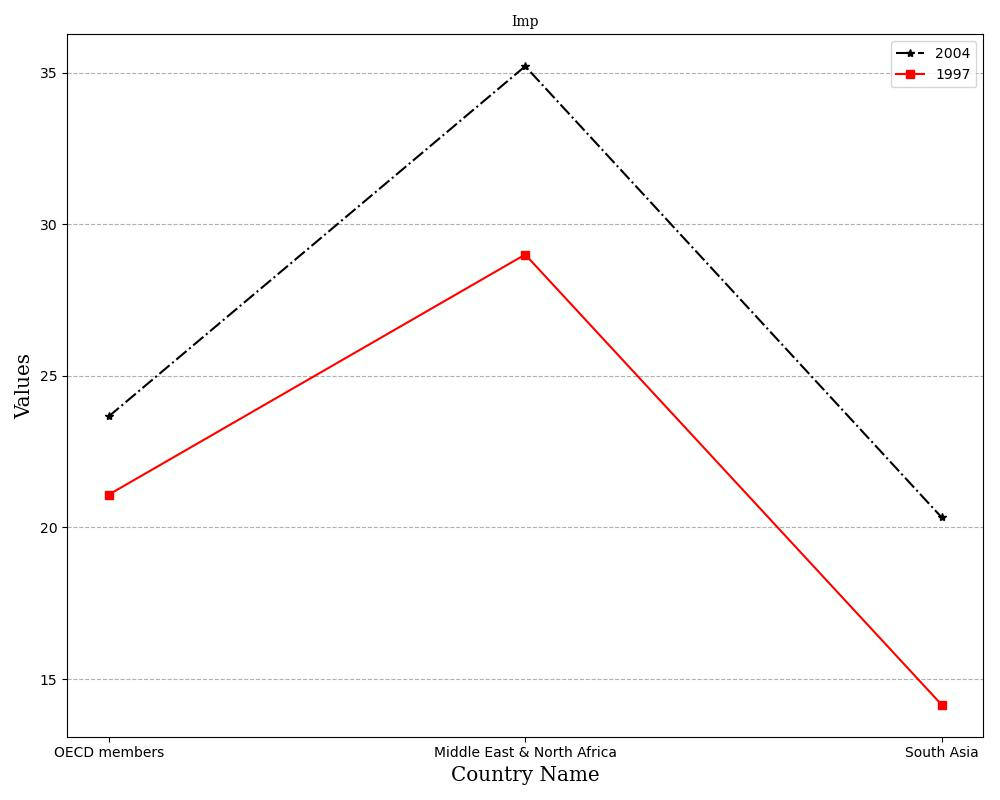}
    \caption[Sample Json]{An example showcasing the JSON configuration for a line chart alongside the generated chart itself. It is important to note that the JSON includes the plotting parameters and the underlying data table. This inclusion ensures a clear and discriminative association between the different data series and their respective visual attributes.}
        \label{fig:json_example}
    \end{minipage}\hfill
    \begin{minipage}{0.45\textwidth}
        \centering
        \begin{verbatim}
{
    ``data_table'': ``Country Name | 2004 | 1997 <0x0A> 
    OECD members | 23.66 | 21.08 <0x0A> 
    Middle East & North Africa | 35.21 | 29.0 
    <0x0A> South Asia | 20.33 | 14.15 <0x0A> '',
    
    ``chart_title'': ``Imp'',
    ``x_axis_title'': ``Country Name'',
    ``y_axis_title'': ``Values'',
    ``global_properties'': {
        ``chart_type'': ``line'',
        ``x_label_params'': {
            ``fontname'': ``Serif'',
            ``fontsize'': ``x-large''
        },
        ``y_label_params'': {
            ``fontname'': ``Serif'',
            ``fontsize'': ``x-large''
        },
        ``legend_params'': {
            ``loc'': 1,
            ``ncol'': 1
        },
        ``chart_title_params'': {
            ``fontname'': ``Serif'',
            ``fontsize'': ``medium'',
            ``rotation'': 0
        },
        ``x_tick_params'': {
            ``axis'': ``x'',
            ``which'': ``major'',
            ``rotation'': 0,
            ``labelsize'': ``medium'',
            ``labelfontfamily'': ``sans-serif''
        },
        ``y_tick_params'': {
            ``axis'': ``y'',
            ``which'': ``major'',
            ``rotation'': 0,
            ``labelsize'': ``medium'',
            ``labelfontfamily'': ``sans-serif''
        },
        ``grid_params'': {
            ``visible'': true,
            ``axis'': ``y'',
            ``linestyle'': ``dashed''
        }
    },
    ``line_properties'': {
        ``linestyles'': [
            ``dashdot'',
            ``solid''
        ],
        ``markers'': [
            ``*'',
            ``s''
        ],
        ``colors'': [
            ``k'',
            ``r''
        ]
    }
}
        \end{verbatim}    
        \end{minipage}
\end{figure}

Fig,\ref{fig:json_example} shows one example of our visual attribute and underlying data JSON format of a line chart. The underlying data consists of three values: `data\_table', 'chart\_title', 'x\_axis\_title', 'y\_axis\_title'. We put the data table in front of the visual attributes since the model's text generation is more stable at front than back, while the data is more sensitive and valuable than visual attributes speaking to real application.\par 

\end{document}